\pdfoutput=1
\documentclass{article}
\PassOptionsToPackage{round}{natbib}

\usepackage[preprint]{single_column}

\usepackage{color}
\definecolor{darkblue}{rgb}{0,0.08,0.45}
\usepackage[colorlinks=true,allcolors=darkblue]{hyperref}       
\usepackage{url}            
\usepackage{booktabs}       
\usepackage{amsfonts}       
\usepackage{nicefrac}       
\usepackage{microtype}      

\renewcommand{\cite}{\citep}
\usepackage{comment}
\usepackage{amsmath}
\usepackage{amsthm}
\usepackage{newtxtext,newtxmath}
\usepackage{braket}
\usepackage{mathtools}
\usepackage[ruled,vlined,linesnumbered]{algorithm2e}
\usepackage{wrapfig}
\usepackage{siunitx}
\usepackage{tabularx}
\usepackage{etoolbox}
\newcommand{\ubold}{\fontseries{b}\selectfont}
\robustify\ubold

\newtheoremstyle{mydefinition}
  {5.5pt} 
  {0pt} 
  {} 
  {} 
  {\bfseries} 
  {.} 
  {.5em} 
  {} 
\newtheoremstyle{mytheorem}
  {5.5pt} 
  {0pt} 
  {\itshape} 
  {} 
  {\bfseries} 
  {.} 
  {.5em} 
  {} 
\theoremstyle{mydefinition}

\theoremstyle{mytheorem}

\newcounter{enum2}
\renewenvironment{enumerate}{%
\begin{list}%
 {\arabic{enum2}.\ \,}{%
 \usecounter{enum2}
 \setlength{\itemindent}{0pt}
 \setlength{\leftmargin}{12pt}
 \setlength{\rightmargin}{0pt}
 \setlength{\labelsep}{0pt}
 \setlength{\labelwidth}{20pt}
 \setlength{\itemsep}{0pt}
 \setlength{\parsep}{0pt}
 \setlength{\listparindent}{0pt}
 \setlength{\topsep}{0pt}
 }}{\end{list}}

\renewcommand{\epsilon}{\varepsilon}
\renewcommand{\phi}{\varphi}
\renewcommand{\log}{\mathop{\textrm{log}}}

\title{Unintended Effects on Adaptive Learning Rate for \\ Training Neural Network with Output Scale Change}

\author{
  Ryuichi Kanoh$^{\mathbf{1,2}}$, \, Mahito Sugiyama$^{\mathbf{1,2}}$  \\
  $^{1}$National Institute of Informatics\\
  $^{2}$The Graduate University for Advanced Studies, SOKENDAI\\
  \texttt{\{kanoh, mahito\}@nii.ac.jp} \\
}

\begin{document}

\maketitle

\begin{abstract}
A multiplicative constant scaling factor is often applied to the model output to adjust the dynamics of neural network parameters. This has been used as one of the key interventions in an empirical study of lazy and active behavior.
However, the present article shows that the combination of such scaling and a commonly used adaptive learning rate optimizer strongly affects the training behavior of the neural network. This is problematic because it can cause \emph{unintended behavior} of neural networks, resulting in the misinterpretation of experimental results. Specifically, for some scaling settings, the effect of the adaptive learning rate disappears or is strongly influenced by the scaling factor. To avoid the unintended effect, we present a modification of an optimization algorithm and demonstrate remarkable differences between adaptive learning rate optimization and simple gradient descent, especially with a small ($<1.0$) scaling factor.
\end{abstract}

\section{Introduction} \label{sec:intro}
Deep learning \cite{lecun2015deeplearning} has penetrated machine learning and data analytics, and is used in a variety of applications. However, its behavior is not well understood.
In recent years, various insights have been achieved by linearly approximating the training of neural networks.
One of the most common tools used with a linear approximation is {\it{Neural Tangent Kernel}} (NTK) \cite{NIPS2018_8076}.
If the change of the NTK during training is negligible, which implies a small parameter change during training, the NTK becomes a vital tool for explaining the good trainability and generalization performance of neural networks~\cite{pmlr-v97-allen-zhu19a, du2018gradient, pmlr-v97-arora19a, NIPS2019_9063}.
However, the linear approximation may not work for realistic neural network models \citep{NEURIPS2019_c133fb1b}.
Therefore, a number of studies \cite{NIPS2019_8559,dAscoli2020DoubleTI,Geiger_2020_disentangle,Geiger_2020, pmlr-v125-woodworth20a} have investigated the difference in behavior between lazy regimes and active regimes.
Here, a \emph{lazy regime} is a regime in which the change in parameters is small relative to the initial value, and a linear approximation is reasonable.
An \emph{active regime}, in contrast, is one regime in which the change in parameters is not small, and a linear approximation is no longer valid.

To empirically compare the behavior of the lazy regime and active regime, \citet{NIPS2019_8559} introduced a useful method, which multiplies the output of the neural network by a positive constant scaling factor $\alpha$,
\begin{equation}
	F( \theta, x ) = \alpha f(\theta, x),
\end{equation}
where $f$ and $F$ are the original and scaled output of a neural network with an input $x$, respectively.
If the scaling factor $\alpha$ is large, even small changes in model parameters can significantly change the output, so the behavior approaches laziness.
In contrast, if $\alpha$ is close to zero, the amount of parameter change is relatively large, and the behavior becomes active. 
Using these properties, \citet{NIPS2019_8559} conducted an empirical study using image recognition models such as ResNet \cite{He_2016_CVPR} and VGG \cite{Simonyan15}, and showed that lazy training does not perform well.
This finding implies that understanding practical neural networks' success requires an understanding outside the framework of linear approximation.
As a result, analyses beyond linear approximation are becoming more widespread \cite{pmlr-v125-li20a, Bai2020Beyond}. For instance, the {\it{mean-field}} (MF) theory is used \cite{MeiE7665,NEURIPS2018_a1afc58c} to describe training dynamics of the neural network without the laziness assumption. Compared to the NTK theory, the values multiplied by the final layer scaling are set to be smaller in the MF theory, resulting in more active behavior. 

While there are a number of examples that demonstrate the importance of active behavior as we described above, this does not necessarily mean that lazy behavior does not benefit.
For example, \citet{Geiger_2020_disentangle} and \citet{Lee2020FiniteVI} showed that the appropriate regime depends on the dataset and neural network architecture, and that lazy training often outperforms active training on fully connected neural networks.
\citet{Arora2020Harnessing} showed that the model trained using the NTK, taking over lazy training, performs better than the standard neural network model on UCI datasets.
\citet{NIPS2019_8809} evaluated training time and showed that there are cases in which the training with the NTK performs better and faster than the algorithm with standard backpropagation.
Their results indicate that lazy training potentially has both practical and theoretical advantages. Therefore, in real-world applications, the scaling factor can be considered as a hyper-parameter for determining behavior, like the learning rate and the neural network model's structure. Understanding of lazy training via the scaling factor has become an exciting research topic from both theoretical and practical viewpoints.

In this paper, we show that the combination of \emph{output scaling} and an \emph{adaptive learning rate optimizer} strongly affects the neural network training behavior, which leads to misinterpretation of empirical investigation.
An adaptive learning rate optimizer, such as {\it{adaptive moment estimation}} (Adam) \cite{DBLP:journals/corr/KingmaB14}
or {\it{root mean square propagation}} (RMSProp) \cite{Tieleman2012},
are often used to achieve fast and stable behavior~\cite{dAscoli2020DoubleTI, Geiger_2020_disentangle}. We demonstrate that such optimizers induce unintended behavior as the optimization algorithm strongly influences the parameter dynamics.
To counteract unintended effects, we propose a modification of the optimization algorithm and show that it can properly adjust for unintended effects.
With this modification, we can properly compare the behavior with simple gradient descent. In our numerical experiment, using the modified optimizer, we observe the behavioral difference between simple gradient descent and adaptive learning rate optimizer.

We summarize our contributios as follows:
\begin{enumerate}
	\item We point out that the combination of scaling and adaptive learning rate optimizer causes unintended behavior, which induces a misinterpretation of empirical investigations. This may change the results of some previous studies. 
	\item To solve the problem, we propose modifying the optimization algorithm and showing that it can properly adjust for unintended effects.
	\item Using the modified optimizer, we observe the behavioral difference between simple gradient descent and an adaptive learning rate optimizer. Especially, under the setting of the scaling factor $\alpha < 1.0$ and hyper-parameters that give accurate classification:
	\begin{enumerate}
	    \item[(a)~] The range of hyper-parameters with the adaptive learning rate optimizer is wider than that with the simple gradient descent, implying higher robustness to hyper-parameter selection.
	    
	    \item[(b)~] The power law that hyper-parameters follow differs between the simple gradient descent and the adaptive learning rate optimizer. For the same scaling factor, the proper learning rate with the adaptive learning rate optimizer becomes larger than that with the simple gradient descent.

	    \item[(c)~] For the adaptive learning rate optimizer, consistency of hidden features during the training is likely to be smaller than that of the simple gradient descent.
	\end{enumerate}
\end{enumerate}

\section{Unintended effects induced by scaling and adaptive learning rate} \label{sec:unintentional}
The objective of training neural networks is to minimize the following error function:
\begin{equation}
	\mathcal{L}(\theta)=\frac{1}{n} \sum_{(x, y) \in \mathcal{T}} \ell\bigl(f(\theta, x), y\bigr),
	\label{eq:standard_loss}
\end{equation}
where $\ell$ is the loss per sample with a ground truth label $y$, and $\mathcal{T}$ is a training dataset with size $n$.
The loss function introduced in \citet{NIPS2019_8559} for experiments with an output scaling factor is given as
\begin{equation}
	\mathcal{L}(\theta)=\frac{1}{\alpha^{2} n} \sum_{(x, y) \in \mathcal{T}} \ell\Bigl(\alpha\bigl(f(\theta, x)-f(\theta_{0}, x)\bigr), y\Bigr),
	\label{eq:scaling}
\end{equation}
where $\theta_0$ is a model parameter at initialization.
Compared to the standard loss function shown in Equation~\eqref{eq:standard_loss}, there are two differences.
First, it forces the model's output to be zero at the start of training by subtracting the initial prediction value.
This modification prevents an immense loss value at the beginning of the training period when $\alpha \gg 1$.
Second, not only the model output scaling with $\alpha$, but also the loss function is scaled by $\alpha^{-2}$.
With loss scaling by $\alpha^{-2}$, a comparison with different scaling factors is valid because of
\begin{equation}
\alpha \dot{f}(\theta, x) = \alpha \nabla_{\theta} f(\theta, x) \dot{\theta} \sim \mathcal{O} (\alpha^{0}),
\end{equation}
where the $\dot{f}$ and $\dot{\theta}$ are time-derivative of $f$ and $\theta$, and a notation $\mathcal{O}$ is the Bachmann–Landau notation.

As for an optimization algorithm, simple gradient descent is formulated as follows:
\begin{equation}
	{\theta}_{t+1} = \theta_t - \eta G_t,
\end{equation}
where $\eta$ and $G_t$ are a learning rate and the gradient of the loss function (e.g., Equation~\eqref{eq:standard_loss} and~\eqref{eq:scaling}) at the $t$-th step, respectively.

Looking at an adaptive learning rate procedure, the algorithm for parameter update used in RMSProp~\cite{Tieleman2012} is given as
\begin{equation}
	{\theta}_{t+1} = \theta_t - \frac{\eta}{\sqrt{v_t}+\epsilon} G_t,
	\label{eq:ada_1}
\end{equation}
where
\begin{equation}
	v_t = \rho v_{t-1} + (1-\rho)G_t^2,
	\label{eq:ada_2}
\end{equation}
$\epsilon$ is a small scaler value for preventing zero division, and $\rho$ is a decay rate for the weight of recent gradient values.
RMSProp is a special case of Adam~\cite{DBLP:journals/corr/KingmaB14}, with the only difference being that there is no momentum term.

\begin{figure}[t]
	\centering
	\includegraphics[width=8.5cm]{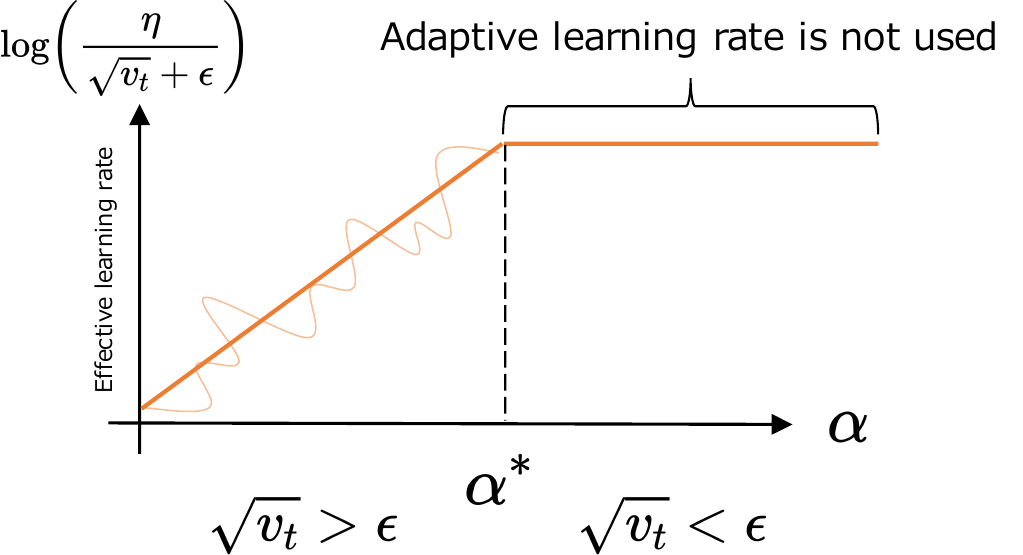}
	\caption{Effective learning rate dependency on the scaling factor with original RMSProp. The wavy line indicates the range in which the adaptive learning rate is used properly. In this case, the effect on adaptive learning rate disappears with $\alpha > \alpha^{\ast}$. Further, even with $\alpha < \alpha^{\ast}$, the effective learning rate depends on $\alpha$, which is not desirable for comparison. }
	\label{fig:schematics1}
\end{figure}

Here we show that combining scaling (Equation~\eqref{eq:scaling}) and an adaptive learning rate algorithm (Equations~\eqref{eq:ada_1} and~\eqref{eq:ada_2}) induces \emph{unintended effects}, while this combination has been already used in some studies (e.g.,~\citet{dAscoli2020DoubleTI, Geiger_2020_disentangle}).
Figure~\ref{fig:schematics1} is the schematic image of the effective learning rate dependency on $\alpha$.
The key observation is that the value of $v_t$ in Equation~\eqref{eq:ada_2}, which is used to adaptively determine the effective learning rate in RMSProp, depends on $\alpha$ as
\begin{equation}
v_t \sim \mathcal{O}(\alpha^{-2}),
\end{equation}
due to the fact that 
\begin{equation}
G_t = {\alpha^{-2}}\nabla_\theta {\mathcal{L}(\alpha f(\theta, x))} \sim \mathcal{O}(\alpha^{-1}).
\end{equation}
Since $\epsilon$ in Equation~\eqref{eq:ada_1} does not depend on $\alpha$, there is a critical value $\alpha^\ast$, where $\sqrt{v_t} = \epsilon$.
If $\alpha$ is sufficiently larger than $\alpha^\ast$, the effect of $v_t$ becomes smaller, and the impact of $\alpha$ on the adaptive learning rate becomes almost negligible.
However, if $\alpha$ is smaller than $\alpha^\ast$, the effect of the adaptive learning rate is present, and the effective learning rate $\eta / (\sqrt{v_t}+\epsilon)$ of RMSProp depends on $\alpha$.
Therefore, \emph{the effect of the output scaling and the effective learning rate change cannot be disentangled}.
Because of this entanglement, we need to update the optimization algorithm to disentangle the effects on change of the scaling factor and the effective learning rate. 

\IncMargin{10pt}
\begin{algorithm}[t]
	\SetKwInOut{Input}{input}
	\SetKwInOut{Initialization}{initialize}
	\SetKwInOut{Output}{output}

	\Input{$\theta_0, x$}
	\Initialization{$v_0 \leftarrow 0$, $t \leftarrow 0$}
	\SetAlgoLined
	\While{$\theta_t$ does not converge}{
		$G_t \leftarrow \frac{1}{\alpha^2} \nabla_\theta \mathcal{L}(\alpha \left(f\left(\theta_t, x \right) - f\left(\theta_0, x)\right)\right)$ \\
		$v_t \leftarrow \rho v_{t} + (1-\rho)(\alpha G_t)^2\quad$ //\ Compared with the original RMSProp, $G_t$ is $\alpha$-folded \\
		${\theta}_{t} \leftarrow \theta_t - \frac{\eta}{\sqrt{v_t}+\epsilon} G_t$ \\
		$t \leftarrow t+1$ \\
	}
	\Output{$\theta_t$}
	\caption{Modified RMSProp optimizer for eliminating scaling factor dependency on learning rate}
	\label{alg:modified_rmsprop}
\end{algorithm}
\DecMargin{10pt}

\begin{figure}[t]
	\centering
	\includegraphics[width=7.5cm]{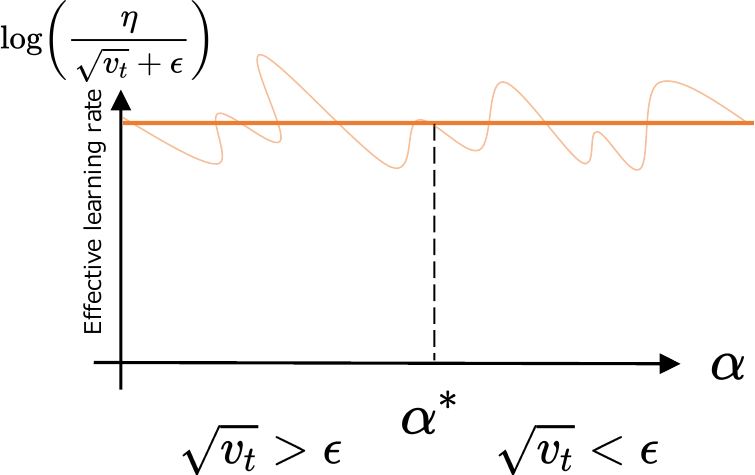}
	\caption{Effective learning rate dependency on the scaling factor with modified RMSProp. The wavy line indicates the range in which the adaptive learning rate is properly used. In this case, the adaptive learning rate's effect does not disappear with $\alpha > \alpha^{\ast}$ and the effective learning rate does not depend on $\alpha$ anywhere.}
	\label{fig:schematics2}
\end{figure}

We present our proposal of the modified RMSProp optimizer in Algorithm~\ref{alg:modified_rmsprop}, which modifies RMSProp to cancel an unintended effect and achieve the disentanglement.
Figure~\ref{fig:schematics2} is the schematic image of the effective learning rate dependency on $\alpha$ with the proposed optimizer.
The difference to Equation~\eqref{eq:ada_2} is that the gradient term $G_t$ used to compute $v_t$ is $\alpha$-folded as,
\begin{equation}
	v_t = \rho v_{t-1} + (1-\rho)(\alpha G_t)^2.
	\label{eq:ada_3}
\end{equation}
It makes $v_t$ independent of $\alpha$ as $G_t \sim \mathcal{O}(\alpha^{-1})$, and the effective learning rate no longer depends on $\alpha$.
Algorithm~\ref{alg:modified_rmsprop} is a generalization of the conventional RMSProp and is consistent with the unmodified behavior when $\alpha=1$.
This modification can be applied not only to RMSProp, but also to other optimization methods, such as Adam, that have the same elements.

\section{Setup on numerical experiments} \label{sec:setup}
We conducted numerical experiments with a two-layer neural network using the modified optimizer in Algorithm~\ref{alg:modified_rmsprop}.
With reference to a similar work, the procedures for the experiments are based on \citet{Geiger_2020_disentangle}.

\paragraph{Model architecture}
A two-layer neural network was used for numerical experiments, which is defined as
\begin{align}
	 \hat{z} &= d^{-\nicefrac{1}{2}} W^0 x,      \\
	 z &= \sigma (\hat{z}),                 \\
	 f(\theta, x) &= h^{-\nicefrac{1}{2}} W^1 z,
\end{align}
where $d \in \mathbb{N}$ is the size of an input vector, and $h \in \mathbb{N}$ is the width of the intermediate layer of the neural network ($1,000$ in numerical experiments).
All of the weights $W^0\in \mathbb{R}$, $W^1\in \mathbb{R}$ are initialized as standard Gaussian random variables, $W^1, W^2 \sim \mathcal{N} (0, 1)$.
For simplicity, the bias parameter was not used.
The scaled softplus function $ \frac{a}{\beta} \ln (1+e^{\beta x})$ was used for activation function $\sigma$,
where $a$ is determined by Monte Carlo method to ensure that the variance of preactivation is $1$. $\beta$ was set it to be 5.

\paragraph{Dataset}
The MNIST~\citep{lecun-mnisthandwrittendigit-2010}, Fashion-MNIST~\citep{xiao2017/online} and CIFAR10~\citep{cifar10} dataset were used for numerical experiments.
Two-dimensional data were converted to a one-dimensional vector (length $d \in \mathbb{N}$) and used as input to a fully connected neural network.
Here, to speed up the experiment, the training dataset was randomly subsampled up to $10,000$ ($20$ percent of the datasets). The dataset for evaluation was not subsampled from the original dataset size ($10,000$ in total).
The input was normalized to be on the sphere $\Sigma_i x_i ^2 = d$.

\paragraph{Loss function}
We used soft hinge loss,
\begin{equation}
\ell\bigl(f(\theta, x), y\bigr) = \frac{1}{\beta} \ln \bigl(1+e^{\beta (1 - f(\theta, x)y)}\bigr),
\end{equation}
for the $10$ class 
classification task,
where $y=\pm 1$ for positive and negative labels for each class, respectively. $\beta=20$ for soft hinge loss.
As described in Section~\ref{sec:unintentional}, loss scaling and initial prediction shift were used to calculate the loss value.

\paragraph{Optimization}
The hyper-parameters of RMSProp, $\rho$ and $\epsilon$ (see Equation~\ref{eq:ada_1}, \ref{eq:ada_2}), were set to $0.999$ and $10^{-8}$, respectively. These values were used for both the modified and unmodified algorithms. $v_0$ was set to be $0$.
We performed $5 \times 10^3$ full-batch gradient descent steps with the simple gradient descent, RMSProp, and modified RMSProp optimizer.
Double precision was used for all calculations. Note that there are some cases where this precision is not sufficient, depending on the setting of $\eta$ and $\alpha$ (Details in Section~\ref{sec:result}).

\paragraph{Metric}
We report top-1 accuracy on the evaluation dataset. Performances for the training dataset are provided in the supplementary material.
We also report a consistency of the hidden features $\hat{z}$ obtained with initial and trained neural network models for evaluating the degree of the dynamics of parameters. Specifically, consistency is the percentage of hidden features on each neuron that have the same sign before and after training.

\section{Result and Discussion} \label{sec:result}

\begin{figure*}
	\centering
	\includegraphics[width=13cm]{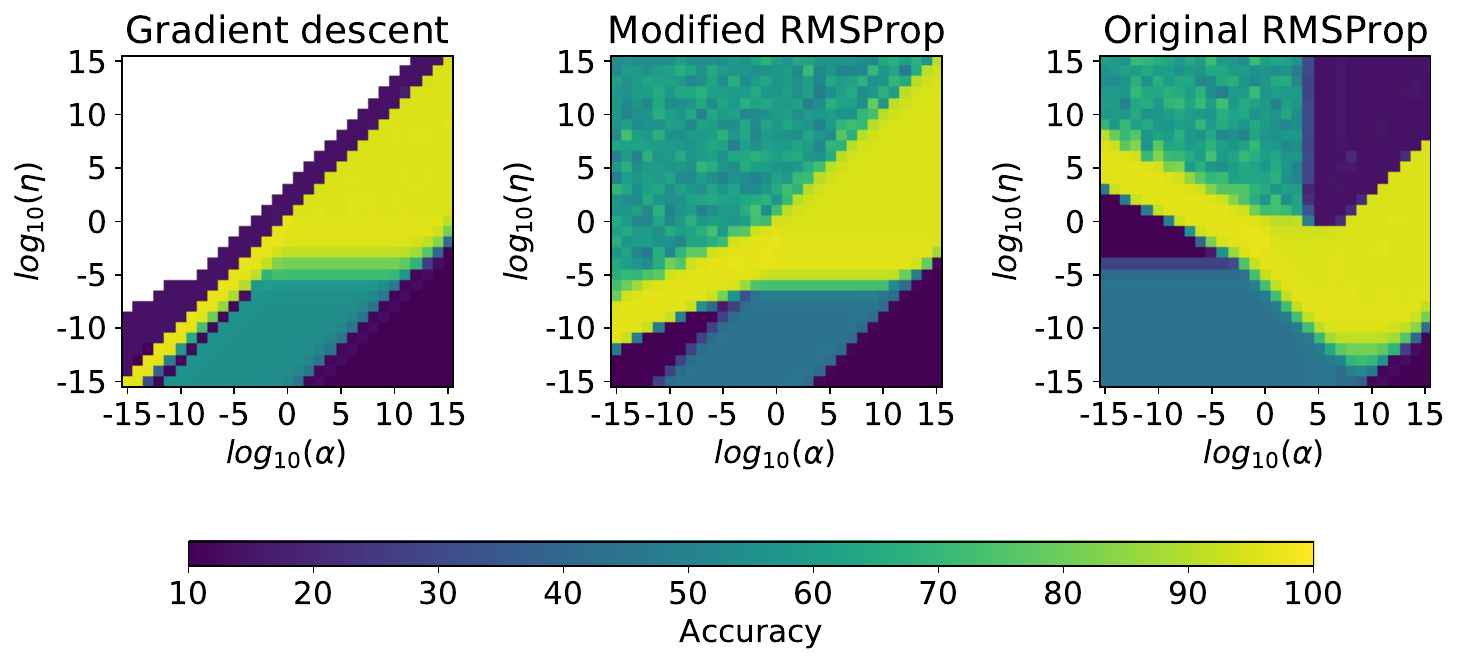}
	\caption{
		Two-dimensional plots of classification accuracy on the MNIST evaluation dataset.
		The white area in the figure of gradient descent represents the divergence of training errors. }
	\label{fig:mnist_acc}
\end{figure*}

\begin{figure*}
	\centering
	\includegraphics[width=13cm]{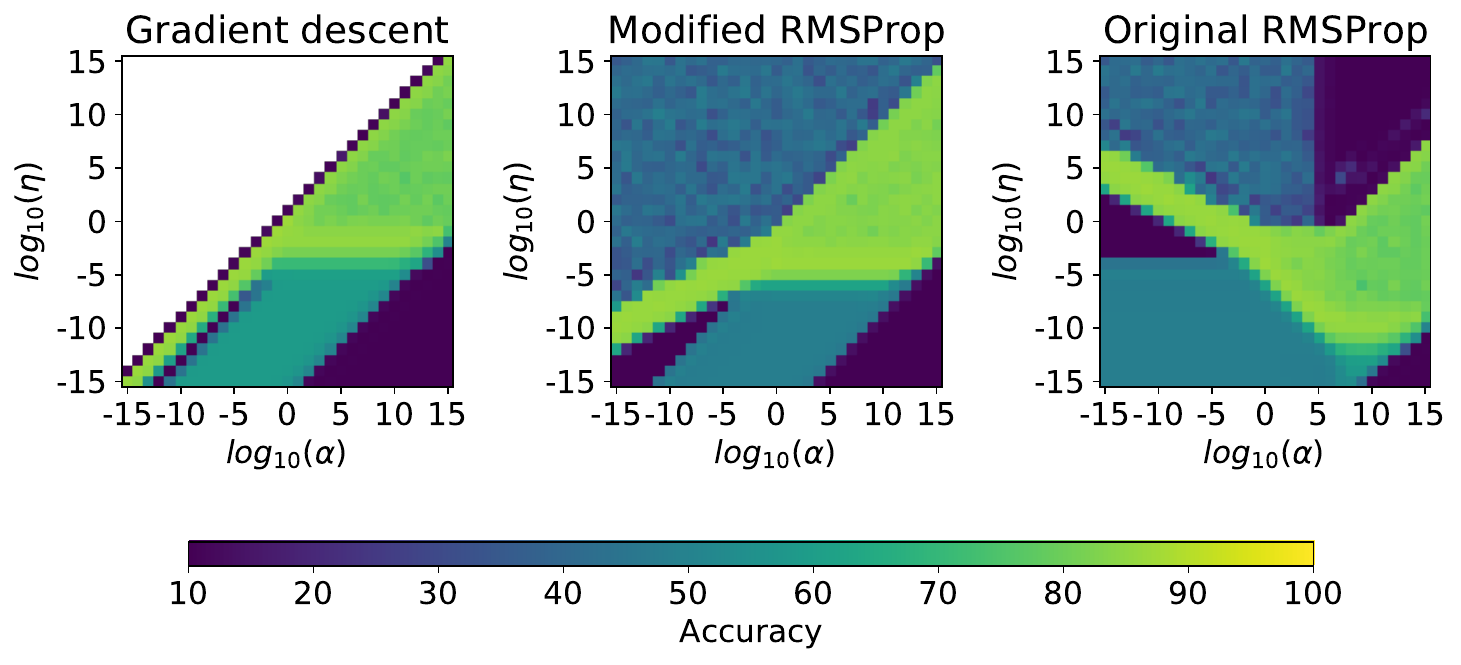}
	\caption{
		Two-dimensional plots of classification accuracy on the Fashion-MNIST evaluation dataset.
		The white area in the figure of gradient descent represents the divergence of training errors. }
	\label{fig:fashion_mnist_acc}
\end{figure*}

\begin{figure*}
	\centering
	\includegraphics[width=13cm]{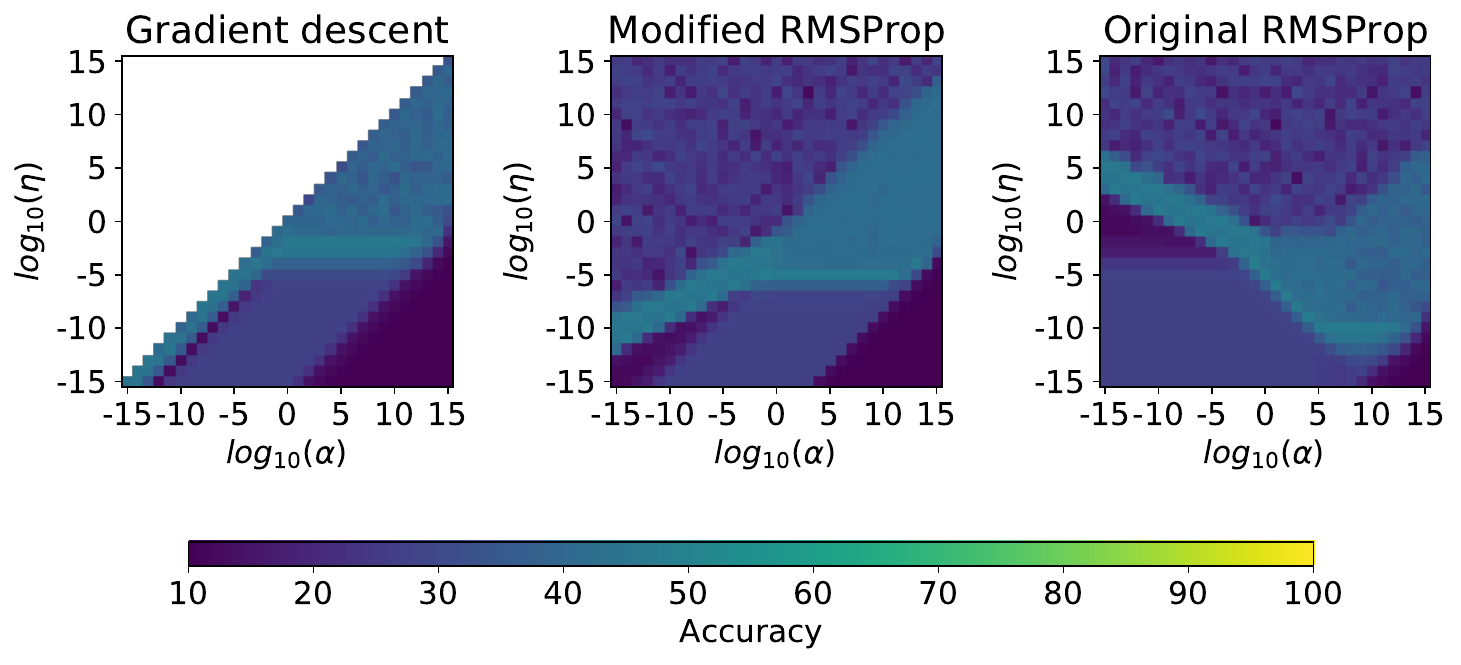}
	\caption{
		Two-dimensional plots of classification accuracy on the CIFAR-10 evaluation dataset.
		The white area in the figure of gradient descent represents the divergence of training errors. }
	\label{fig:cifar10_acc}
\end{figure*}

\subsection{Impacts on the Algorithm Modification}
Figures~\ref{fig:mnist_acc}, \ref{fig:fashion_mnist_acc}, and \ref{fig:cifar10_acc} show classification accuracy on the different datasets with a variety of $\eta$ and $\alpha$, where a significant difference is observed between the modified and the original RMSProp.
On the right-hand side of each plot, implying $\alpha > \alpha^\ast$, the positive trend observed in $\alpha$ and $\eta$ is similar to that of the simple gradient descent and the original RMSProp.
On the left-hand side of each plot, implying $\alpha < \alpha^\ast$, the characteristic slope of the original RMSProp is bent to around $90$ degrees compared to the modified RMSProp.
These observations are consistent with the explanation of the change in the effective learning rate with $\alpha$, described in Section~\ref{sec:unintentional}. It means that our proposed method succeeds in erasing the dependence of effective learning rate on $\alpha$ as intended.
We find that the trend does not change significantly as the dataset changes. 

The positive trend observed in Figures~\ref{fig:mnist_acc}, \ref{fig:fashion_mnist_acc}, and \ref{fig:cifar10_acc} in $\alpha$ and $\eta$ for simple gradient descent and modified RMSProp can be understood by considering the impact of $\alpha$ on the Hessian.
It is known that there is a necessary condition of the learning rate and the eigenvalue of the Hessian for the simple gradient descent training convergence \citep{lecun-98b, NIPS1992_589}, $\eta< 2 / \lambda_{\max}$, where $\lambda_{\max}$ is the maximum eigenvalue of the Hessian.
Since model output is scaled by $\alpha$, its corresponding hessian is also scaled by $\alpha$.
Further, since the loss is scaled by $\alpha^{-2}$ (Equation~\eqref{eq:scaling}) in our experiment, the proper $\eta$ for training that are roughly proportional to $\eta< 2 / \lambda_{\max}$ is scaled by $\mathcal{O}\left(\alpha^{1}\right)$. Therefore, there is a positive trend between $\alpha$ and $\eta$.

Note that performance becomes worse in the region of small $\eta$ and large $\alpha$ because of the lack of numerical precision in computation.
In such a situation, parameter change, $\eta G_t$, becomes relatively small compared to the model parameter $\theta$.
Because of that, we observe a completely unchanged $\theta$ from initialization in our numerical experiments.
Therefore, experiments with such extreme settings should not be trusted.

\begin{figure}
	\centering
	\includegraphics[width=8cm]{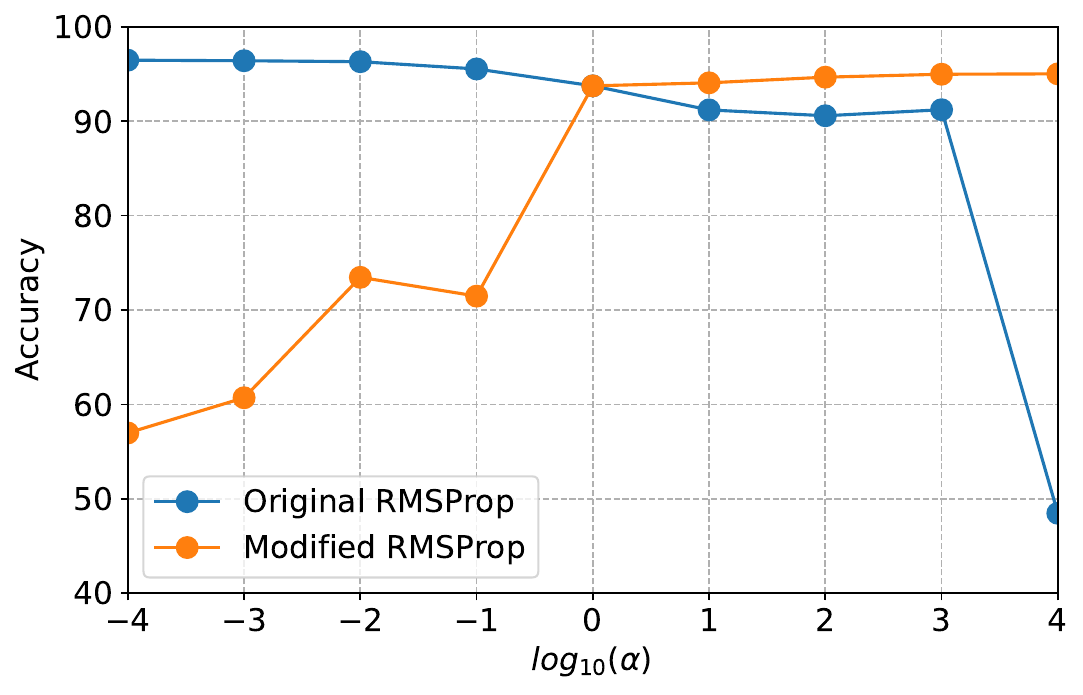}
	\caption{
		Classification accuracy on the MNIST dataset with $\eta=1.0$.	In the case of original RMSProp, the performance is better when $\alpha$ is small, implying active training.
		On the other hand, in modified RMSProp, the performance is better when $\alpha$ is large, implying lazy training.
	}
	\label{fig:1d_acc}
\end{figure}

Figure~\ref{fig:1d_acc} illustrates the classification accuracy with respect to the changes of the scaling factor $\alpha$ of RMSProp with $\eta = 1.0$ fixed before and after modification. The range of $\alpha$ is from $10^{-4} $ to $10^4$ to guarantee sufficient computational precision with $\eta=1.0$.
As the figure shows, the conclusions switch entirely before and after the modification; that is, a larger $\alpha$ (lazy regime) is better for the modified RMSProp and a smaller $\alpha$ (active regime) is better for the original RMSProp.
This observation implies that our modification (Algorithm~\ref{alg:modified_rmsprop}) is crucial in the evaluation of the scaling factor. Additionally, this kind of experimental protocols, which examines performance by changing $\alpha$, has been commonly used in previous studies. Therefore, part of the empirical results \citep{Geiger_2020_disentangle, dAscoli2020DoubleTI} may be affected by these unintended effects.

\subsection{Comparison to the simple gradient descent}

The comparison between the modified RMSProp and a simple gradient descent provides some interesting behavioral changes. Significant differences are often observed when $\alpha$ is small, implying active training.

\subsubsection{Performance robustness}
When we use an adaptive learning rate optimizer with a small $\alpha$, Figures~\ref{fig:mnist_acc}, \ref{fig:fashion_mnist_acc}, and \ref{fig:cifar10_acc} show that there is a wide range of $\eta$ values to achieve high-performance ($\sim 100\%$ accuracy) compared with a simple gradient descent. This implies the effects on the robustness of the performance to the choice of hyper-parameters, and may be an example of the remarkable effect of the adaptive learning rate. 

\subsubsection{Proper hyper-parameter setting}

With the modified RMSProp, the characteristic slope is folded around $\log_{10} \alpha \sim -2$ in Figures~\ref{fig:mnist_acc}, \ref{fig:fashion_mnist_acc}, and \ref{fig:cifar10_acc}. Such folding is not observed with a simple gradient descent case. Because of this behavior, when we use an adaptive learning rate optimizer with a small $\alpha$, the value of an appropriate $\eta$ is likely to be larger than that with a simple gradient descent. Although the reason for such folding is not clear so far, the effect may be due to the switching of inductive biases tied to changes in scaling \citep{pmlr-v125-chizat20a}. For example, the so-called the NTK-regime and MF-regime are known to transition between an output scaling of $1 / \sqrt{h}$ and $1 / h$ \cite{NIPS2018_8076, MeiE7665}, where $h$ is the width of the hidden layer of the model, and these values are similar to where the observed trend transitions occur. Note that $1,000$ is used for $h$ in our numerical experiments (Section~\ref{sec:setup}).

\begin{figure*}
	\centering
	\includegraphics[width=14cm]{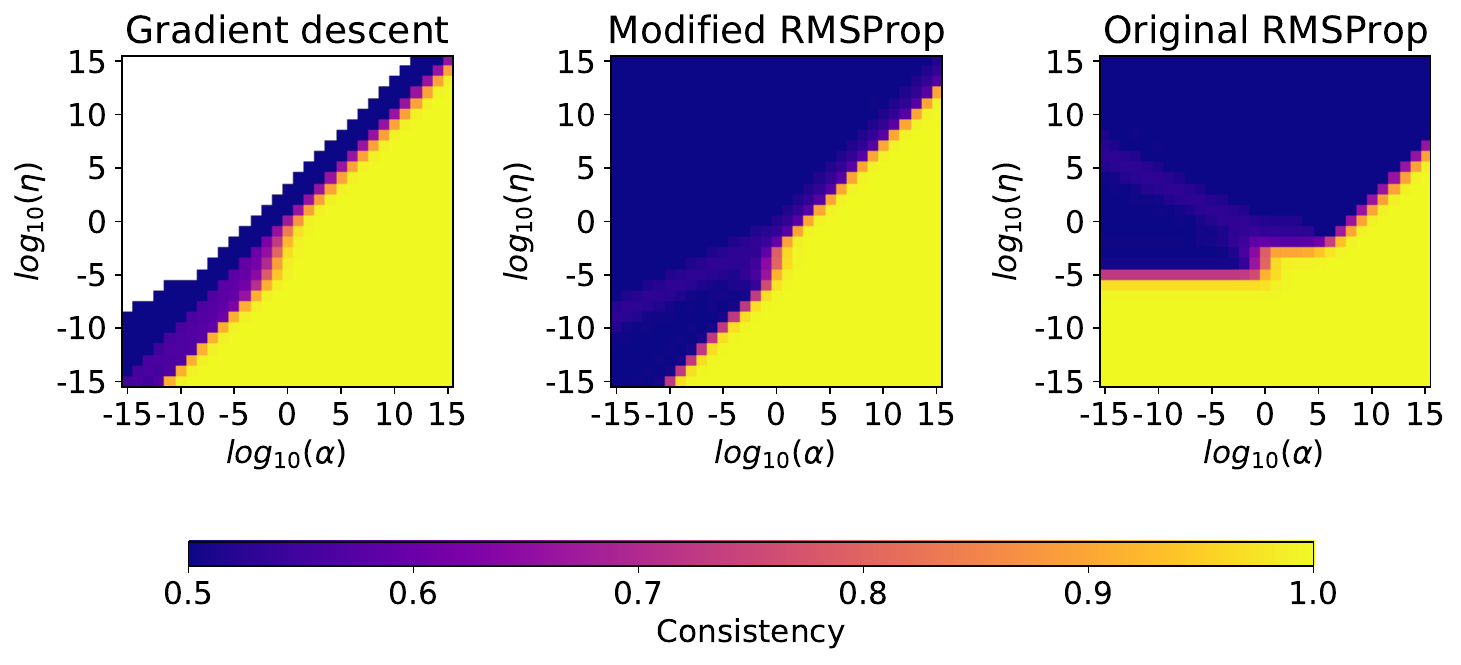}
	\caption{
		Consistency of the hidden feature signs ($\pm$ of $\hat{z}$ for each neuron) obtained with initial and trained neural network models. Hidden features are extracted by use of the evaluation dataset of the MNIST.}
	\label{fig:mnist_consistency}
\end{figure*}

\begin{figure*}
	\centering
	\includegraphics[width=14cm]{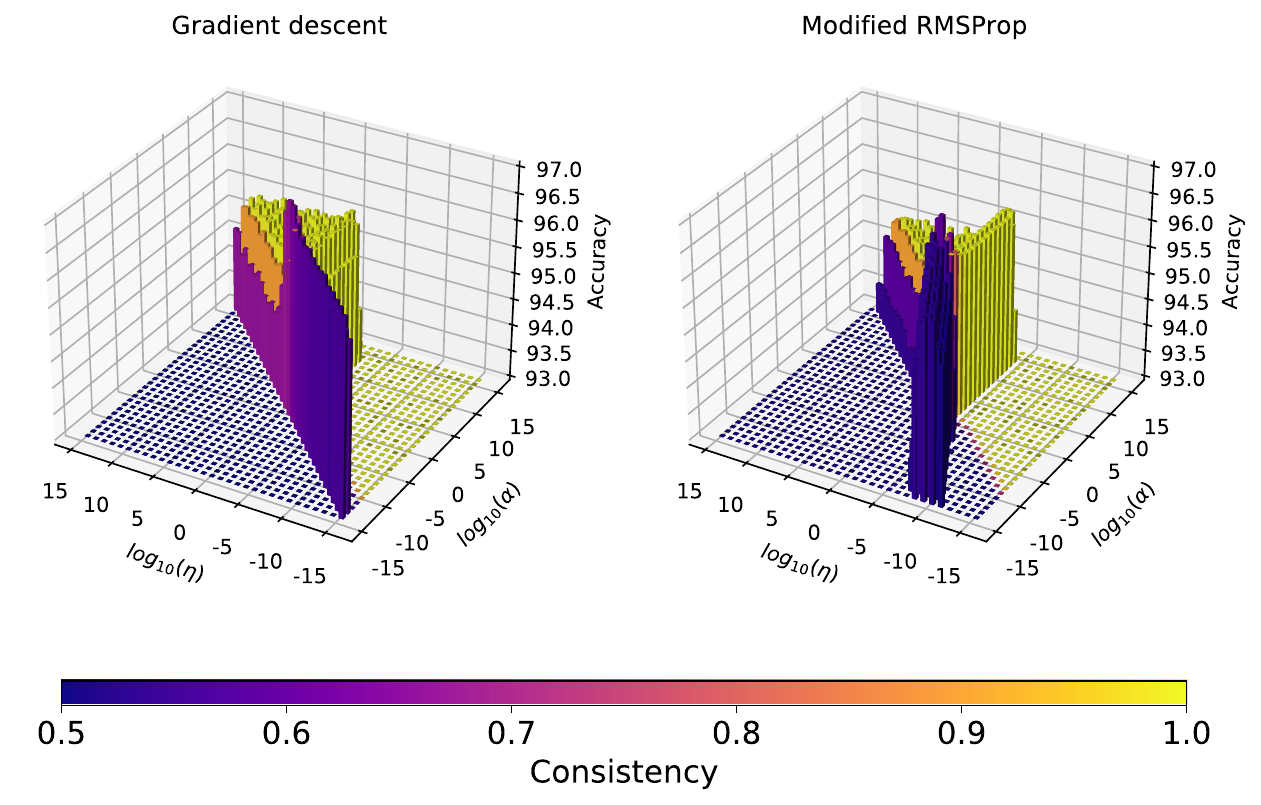}
	\caption{
		Three-dimensional plot that combines Figure~\ref{fig:mnist_acc} and \ref{fig:mnist_consistency}. The color represents the consistency of the hidden feature signs shown in Figure~\ref{fig:mnist_consistency}, and the height of the bar represents the classification accuracy shown in Figure~\ref{fig:mnist_acc}. The vertical and horizontal axes represent the logarithmically scaled learning rate $\eta$ and the scaling factor $\alpha$, as in Figures~\ref{fig:mnist_acc} and \ref{fig:mnist_consistency}. We can see that the accuracy is high in the region where the learning rate and scaling factor are both small. Additionally, the sign consistency of hidden features is small in these regions, indicating that the learning is not lazy and rather has an active behavior.}
	\label{fig:3d}
\end{figure*}

\subsubsection{Hidden feature consistency}

Figure~\ref{fig:mnist_consistency} shows the hidden feature consistency introduced in Section~\ref{sec:setup}.
Figure~\ref{fig:3d} shows the three dimensional plot that combines Figures~\ref{fig:mnist_acc} and \ref{fig:mnist_consistency}. For both simple gradient descent and the modified RMSProp, we can see that the classification accuracy is high in the region where the $\eta$ and $\alpha$ are both small. In addition, the sign consistency of hidden features is small in these regions, indicating that the training behavior is not lazy, but has an active behavior. 

\begin{figure}
	\centering
	\includegraphics[width=8cm]{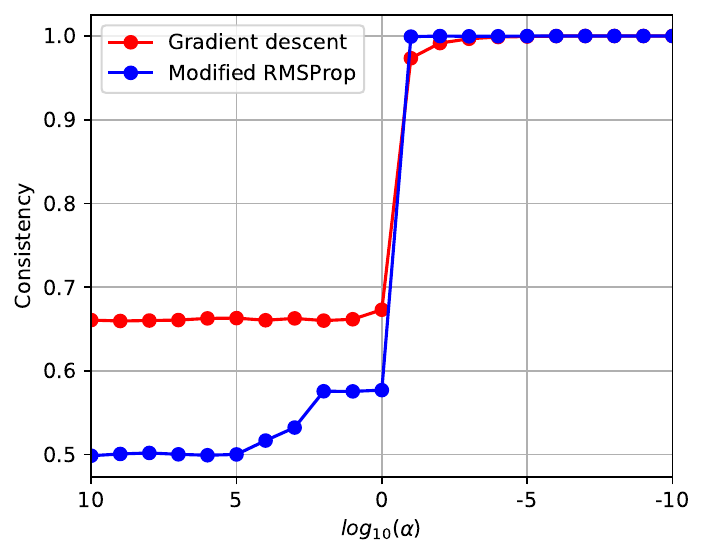}
	\caption{
		Consistency of the hidden feature signs. The learning rate is selected as it has the highest prediction performance for each $\alpha$. Therefore, it should be noted that the value of the learning rate is different for each $\alpha$. Hidden features are extracted by use of the evaluation dataset of the MNIST.
	}
	\label{fig:1d_consistency}
\end{figure}

Careful comparison of the regions with small $\eta$ and $\alpha$ shows that a difference exists between simple gradient descent and the modified RMSProp. Figure~\ref{fig:1d_consistency} shows that the consistency of the hidden feature signs is different when the high-performing hyper-parameters are used. In the region where a small $\alpha$ is used, the change from the initial value is larger when an adaptive learning rate optimizer is used. This means that features are likely to be extracted actively, which may be important when building machine learning models with high interpretability such as the attention mechanism \citep{NIPS2017_3f5ee243}.

\section{Related Work} \label{sec:related}
From the viewpoint of the implications for the deep learning theory, it is important to note that even in previous studies that did not use an adaptive learning rate, the numerical experiments' interpretation can change significantly with different settings of $\eta$.
Some previous studies \citep{NIPS2019_8559, Geiger_2020_disentangle, dAscoli2020DoubleTI} check the performance dependency on $\alpha$ with a single $\eta$ setting. For example, \citet{NIPS2019_8559} reported that performance with $\alpha$ just close to the training diverges is the highest, and a sharp performance decrease as $\alpha$ increases. Such behavior can be reproduced by observing slices at $\eta \sim 10^{-5}$ in our experiment (the gradient descent case in Figures~\ref{fig:mnist_acc},\ref{fig:fashion_mnist_acc}, and \ref{fig:cifar10_acc}). However, such a sharp drop is not observed at larger $\eta$.
It is not desirable to have a situation where theoretical research implications can change drastically with just a small setting change. Ideally, it would be possible to experiment numerically with a continuous gradient flow rather than a gradient method of accumulating discrete steps, but that technique has not been established yet. It may be worthwhile deepening this direction of research \citep{Geiger_2020_disentangle}.

One possible reason for good performance observed near the boundary of the training divergence observed in \citet{NIPS2019_8559} and Figures~\ref{fig:mnist_acc},\ref{fig:fashion_mnist_acc}, and \ref{fig:cifar10_acc} is the existence of the so-called {\it{Catapult phase}}. \citet{Lewkowycz2020TheLL} showed there are three training dynamics ({\it{lazy phase}}, {\it{catapult phase}}, {\it{divergent phase}}) based on the learning rate setting. Observed good performance and learning rate configuration looks to be consistent with the behavior of the {\it{catapult phase}}. For the modified RMSProp, although the training has not diverged numerically as it did in the simple gradient descent cases, a similar trend has been observed.

\section{Conclusion}
In this paper, we have shown that when output scale change and an adaptive learning rate optimizer are used simultaneously in training a neural network, the effective learning rate unintentionally depends on the scaling factor $\alpha$. Such behavior can lead to the misinterpretation of experimental results. Therefore, we have proposed an optimizer for canceling the dependency of the $\alpha$ on the effective learning rate.
Using the modified optimizer, we have succeeded, for the first time, in comparing the changes in behavior with and without adaptive learning rate. We have observed some interesting phenomena, especially with $\alpha < 1.0$, such as changes in robustness, proper hyper-parameter setting, and hidden feature consistency during the training.

\section*{Acknowledgement}
This work was supported by JST, PRESTO Grant Number JPMJPR1855, Japan and JSPS KAKENHI Grant Number JP21H03503 (MS).

\bibliography{references}

\appendix
\section{Performances on training dataset} \label{app:train}
\begin{figure}
	\centering
	\includegraphics[width=13cm]{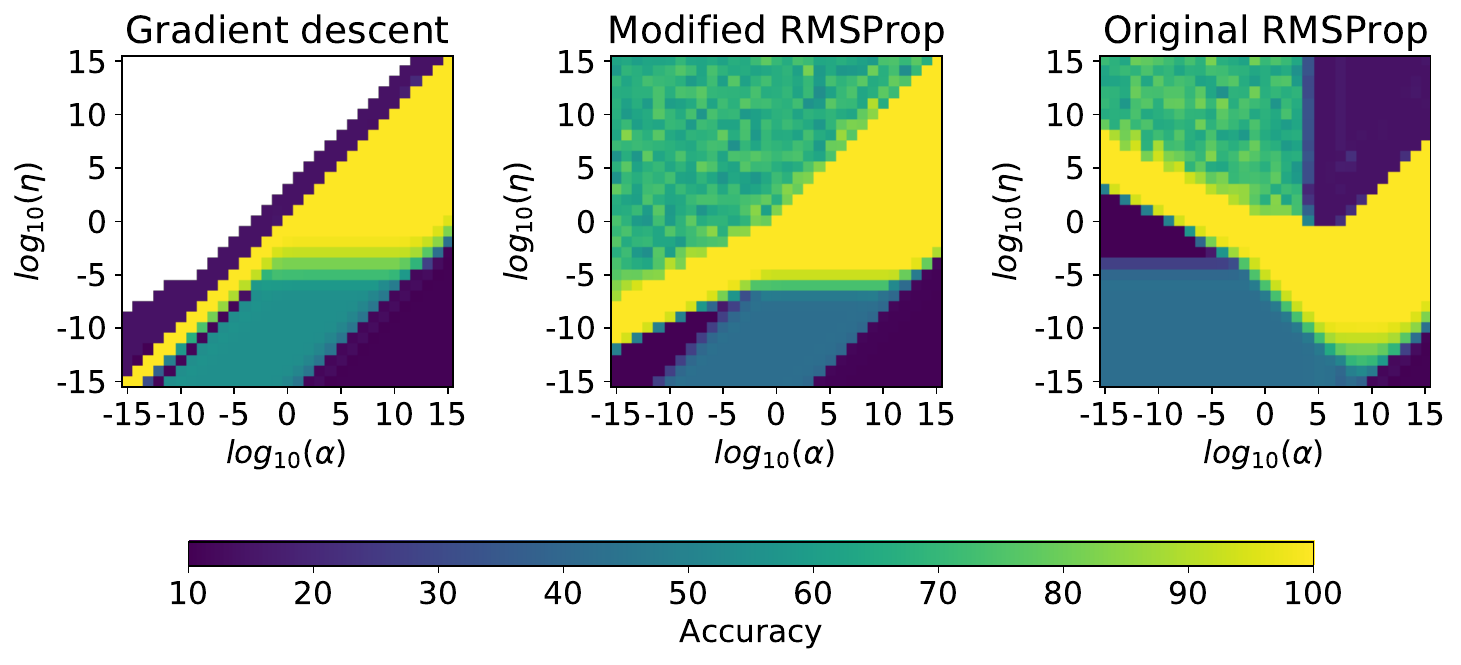}
	\caption{
		Two-dimensional plots of classification accuracy on the MNIST training dataset.
		The white area in the figure of gradient descent represents the divergence of training errors. }
	\label{fig:mnist_acc_tr}
\end{figure}

\begin{figure}
	\centering
	\includegraphics[width=13cm]{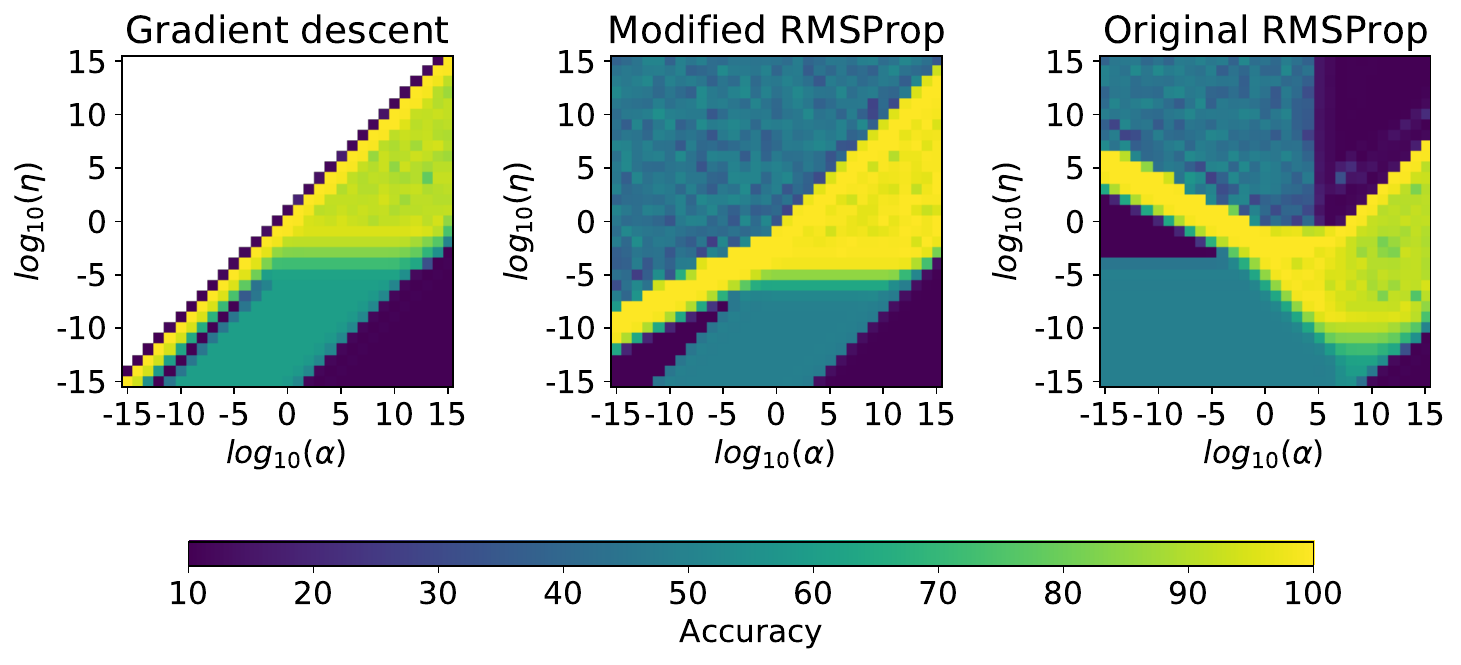}
	\caption{
		Two-dimensional plots of classification accuracy on the Fashion-MNIST training dataset.
		The white area in the figure of gradient descent represents the divergence of training errors. }
	\label{fig:fashion_mnist_acc_tr}
\end{figure}

\begin{figure}
	\centering
	\includegraphics[width=13cm]{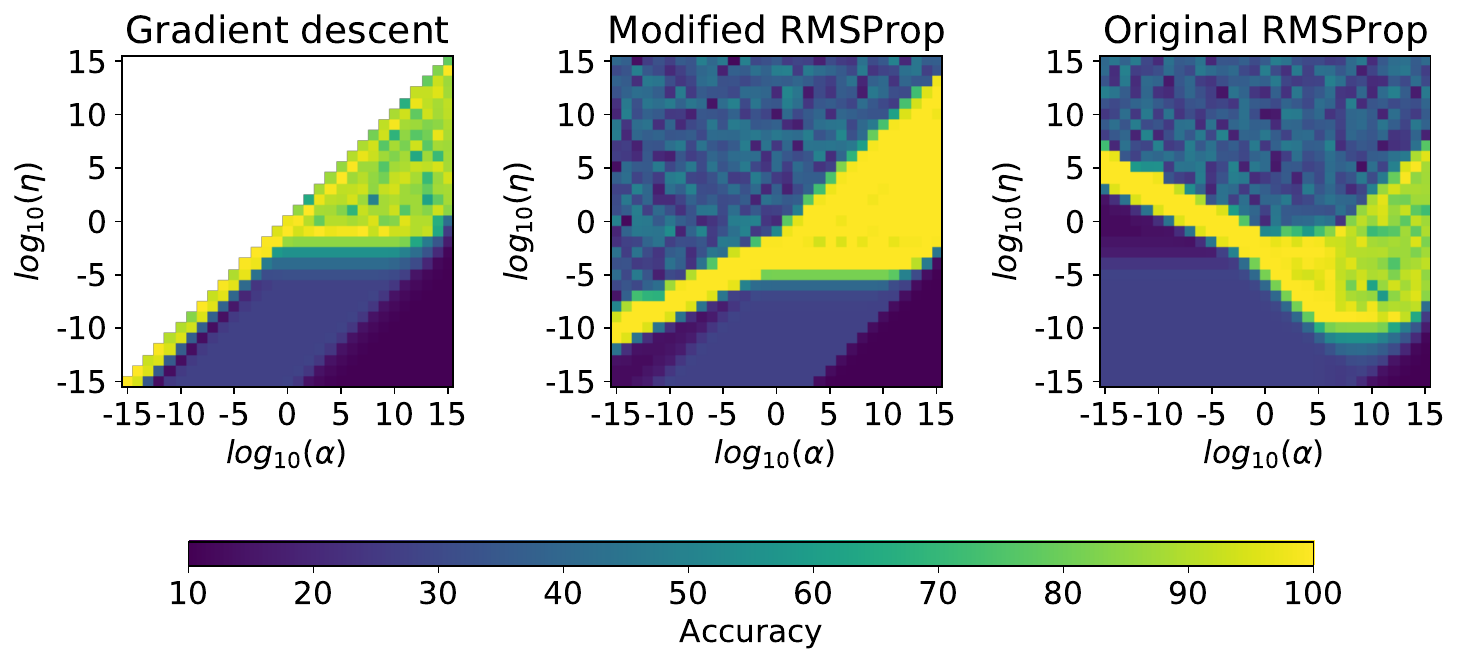}
	\caption{
		Two-dimensional plots of classification accuracy on the CIFAR-10 training dataset.
		The white area in the figure of gradient descent represents the divergence of training errors. }
	\label{fig:cifar10_acc_tr}
\end{figure}

In addition to the performance for evaluation dataset shown in Figures~\ref{fig:mnist_acc}, \ref{fig:fashion_mnist_acc} and \ref{fig:cifar10_acc}, we provide the performance for training dataset in Figures~\ref{fig:mnist_acc_tr}, \ref{fig:fashion_mnist_acc_tr} and \ref{fig:cifar10_acc_tr}. 

\newpage

\end{document}